\documentclass[letterpaper,12pt]{article}
\usepackage[utf8]{inputenc}
\usepackage[margin=1in]{geometry}
\usepackage[normalem]{ulem}

\usepackage[notes,backend=biber]{biblatex-chicago}
\bibliography{darwin,refs,software,cognition}

\usepackage{enumitem}
\setlist{nosep}

\usepackage{setspace}
\usepackage{graphicx}

\newcommand{\citet}{\autocite}
\newcommand{\citep}{\autocite}

\begin{document}
\title{The Development of Darwin's Origin of Species}
\author{Jaimie Murdock, Colin Allen, and Simon DeDeo}
\date{February 24, 2018}
\maketitle

\bigskip

%First, we trained a 200-topic model of Darwin's readings and writings, based on the same corpus as \citet{Murdock2017} (see \ref{sec:lda}). Then, we took 1100 samples for each of the five texts in the study using the LDA query sampling method in \ref{sec:query-sampling}. For each group of samples, we applied k-means clustering to the topic distributions, using the Jensen-Shannon Distance (see \ref{sec:jsd}) between points. We selected the number of clusters based on a silhouette analysis, with a threshold value of 0.5, under which no clusters were assigned, assuming a non-discriminable population (see \ref{sec:silhouette}).

%\onehalfspacing
%\section{Introduction}

From 1837, when he returned to England aboard the \emph{HMS Beagle}, to 1860, just after publication of \emph{The Origin of Species}, Charles Darwin kept detailed notes of each book he read or wanted to  read. His notes and manuscripts provide information about decades of individual scientific practice. The textual nature of these records make them particularly amenable to computational analysis. 

Previously\citep{Murdock2017}, we located digitized versions of the English-language, non-fiction works listed as read in Darwin's reading notebooks. We trained topic models on the full texts of each reading, without using any information about his publications or additional historical metadata. We applied information-theoretic measures to the topic models of Darwin's readings to detect that changes in his reading patterns coincided with the boundaries of his three major intellectual projects in the period 1837-1860: a first shift in 1846 from his \emph{Beagle} research journals to describing the \emph{Cirripedia} --- his ``beloved barnacles’' --- and a second shift in 1854 from barnacles to the work leading to \emph{The Origin of Species} in 1859. While these intellectual phases are well-known to historians, the model offered a novel contribution by highlighting what changed in Darwin's reading habits: as he prepared his notes for the ``large work on species,'' his readings became more exploratory; \emph{i.e.}, as he was organizing his species notes his readings shifted subjects more often than when assembling his \emph{Beagle} and \emph{Cirripedia} notes. %--- as defined by our measure of ``cognitive surprise'', the Kullback-Leibler (KL) divergence (see Methods, below).

%Topic models can also be seen as a (partial) theory of writing, where new documents are generated by drawing words from the topics. 
In this new work we apply the reading model to  five additional documents, four of them by Darwin: the first edition of \emph{The Origin of Species}, two private essays stating intermediate forms of his theory in 1842 and 1844, a third essay of disputed dating, either preceding or succeeding the 1842 essay, and Alfred Russel Wallace's essay, which Darwin received in 1858. After describing the models and what they can show about the relationship between Darwin's writings and his readings over the course of his investigations, we address three historical inquiries, previously treated qualitatively:

\begin{enumerate}
\item the mythology of ``Darwin's Delay,'' that despite completing an extensive draft in 1844, Darwin waited until 1859 to publish \emph{The Origin of Species} due to external pressures  \citep{Gruber1974,Browne2006,VanWyhe2007,richards2015}; 
\item the relationship between Darwin and Wallace's contemporaneous theories, especially in light of their joint presentation \citep{Merton1957,VanWyhe2013,Costa2014}; and
\item the dating of the ``Outline and Draft'' which was rediscovered in 1975 and postulated first as an 1839 draft preceding the Sketch of 1842 \citep{Vorzimmer1975}, then as an interstitial draft between the 1842 and 1844 essays \citep{Schweber1977,Kohn1982}.
\end{enumerate}

\section{Query Sampling the Writings}

Our starting point for answering these questions was the previously trained topic model of his readings, as we were interested in how the readings influenced the writings. Topic models represent each text as a blend of different topics, with each topic being a probability distribution over the words in the collection. The models are statistically derived from a set of texts through joint inference of their word-topic and topic-document distributions \citep{Blei2012}.  Query sampling allows a mixture of topics from a prior model to be assigned to documents not in the original training set. An initially random assignment of the words to the topics is revised iteratively until the assignment stabilizes using the same method used to train the original model. Because of the random starting point, running the query sampling process multiple times leads to different topic distributions for the same text.

This variability in outputs is something to be understood and harnessed, not feared, supporting different perspectives on the text.\citep{Rockwell2016,Allen2017} For any text, there is not claimed to be a single ``correct'' interpretation but rather a set of interpretations in dialog with one another. Digital methods can augment existing debates in the humanities by providing different ways of looking at the text.  

\begin{figure}
\begin{center}
\includegraphics[width=\textwidth]{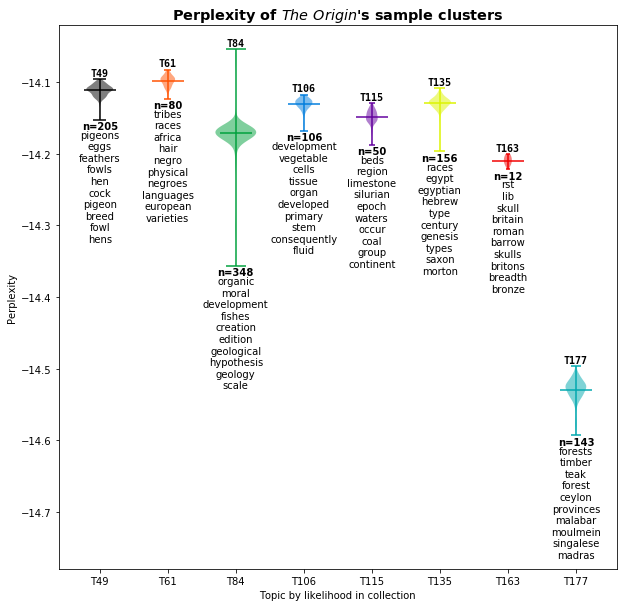}
\end{center}
\caption{\textbf{Cluster Analysis of \emph{The Origin of Species}} --- This ``violin'' plot shows the distribution of perplexity (fit to the document) by topic cluster for \emph{The Origin of Species}. The number below each cluster shows the number of samples classified in that group, and the surface area is proportional to this number. The horizontal line in the center of each violin shows the median perplexity, while the vertical lines span the outliers in each cluster. }
\label{fig:origin-clusters}
\end{figure}

We  approach the diversity of the sampled results by applying a clustering algorithm to the topic distributions, using the silhouette method to choose the number of clusters\citep{Rousseeuw1987}. For \emph{The Origin}, this method detects eight clusters, shown in Figure~\ref{fig:origin-clusters}. Each cluster has a different highest-probability topic. These dominant topics characterize the primary interpretation of the text for each cluster. Inspection of the topics reveals that they are immediately applicable to \emph{The Origin}. For example, pigeons (T49) provide a significant example for Darwin. The dominant topic of the largest cluster (T84) captures some key theoretic concerns with the words `'`development'', ``creation'', ``geological'', ``organic''.

%As historians and philosophers of science we are interested in the utility of computational methods to answer  historical and philosophical questions.  As DH scholars we are challenged to find ways to convey the results of these analyses to non-technical readers and to establish their robustness despite the variation.
%In our analysis, each fitness peak in the topic landscape represents a particular distribution of topics, which can be treated as a way to interpret the text. % TODO: say more about this without entirely elucidating fitness landscapes.

Because the topics fit to \emph{The Origin} by query sampling are derived from the model of the readings, some of the words that have a high probability for a topic in the readings are likely not to appear in \emph{The Origin} at all. For example, the second most likely word in T84, ``moral'' does not appear in the first edition of \emph{The Origin}. Likewise, some of the geographic terms prominent in T177 do not appear in the book.  Indeed, T177 (with terms related to forests and South Asian geography and culture) presents an idiosyncratic view of \emph{The Origin}. The statistical ``perplexity'' of this cluster with respect to the text confirms a relatively poor fit. Nonetheless, the assignment of T177 is grounded both in Darwin's reading of Falconer's \emph{Report on the teak forests of the Tenasserim provinces} in 1853, and in his writing--Falconer is mentioned six times in the first edition of \emph{The Origin}, and related issues are discussed in passages such as this, from chapter 5: 
% * <prof.colin.allen@gmail.com> 2018-02-24T15:54:27.104Z:
% 
% > in 1853
% For later expansion note that this corresponds to a discontinuity in the KL divergence chart that can be attributed to T177/Falconer
% 
% ^.
\begin{quote}[W]e have evidence, in the case of some few plants, of their becoming, to a certain extent, naturally habituated to different temperatures, or becoming acclimatised: ... trees growing at different heights on the Himalaya, were found in this country to possess different constitutional powers of resisting cold. Mr. Thwaites informs me that he has observed similar facts in Ceylon.
\end{quote}
T177, like other clusters featuring geographical and ethnographic terms (T61, T135, T163), highlights how Darwin's own travels, correspondence with other travelers, and reading their published accounts expanded the global range of his evidence.

\section{Measuring Cognitive Surprise}
%In answering all three questions, we make quantitative comparisons between various texts to make our argument: Darwin's writings (in particular, the ``Pencil Sketch'', \emph{Sketch of 1842}, \emph{Draft of 1844}, and the first edition of \emph{The Origin}), Wallace's 1858 manuscript, and the books Darwin read. We use topic modeling to extract the major themes from these texts.
We compared the writings to the readings and each other using an information-theoretic measure of cognitive surprise -- Kullback-Leibler (KL) divergence\citep{Kullback1951} --  used in our previous study and which has proven successful in various cognitive science applications\citep{Murdock2017}.  Applied to the topic distributions derived by query sampling, KL divergence measures the extent to which the distribution of topics encountered in a new text violate the expectations based on the topic distributions in previously encountered texts.

KL divergence is an asymmetric measure, meaning that encountering B after A may generate a different amount of surprise than encountering A after B. Asymmetric measures are useful in many contexts: for example, travel time may be the more useful measure if it will take longer to climb a mountain than to go down it, even though the distance traveled in kilometers is the same. When a symmetric measure of distance between volumes is more appropriate, we use the symmetrical Jensen-Shannon distance (JSD), which is derived from the KL divergence and satisfies the mathematical properties of a distance metric.\citep{lin1991divergence,nielsen2010family,fuglede2004jensen}

\section{Findings}
\subsection{Explaining Darwin's Delay}
Darwin began drafting his theory long before he started organizing his notes in 1854. With two private essays written in 1842 and 1844, it is a historical curiosity that he would wait until 1859 to publish his work, especially as immediately after finishing the second essay he wrote to his wife, Emma, with an addendum to his will concerning publication instructions should he die before finishing his work \citet{DCP761}. This period has become known as ``Darwin's Delay'' \citep{VanWyhe2007}. Theories about its cause include general fear of persecution \citet{Gruber1974}, the anonymous 1844 publication of \emph{Vestiges of the Natural History of Creation}\citet{Chambers1844} highlighting gaps in Darwin's argument\citet{Browne1995}, and extended illness \citep{Gruber1974,VanWyhe2007,richards2015}. We provide evidence for another motivation for the delay that has been proposed by others\citet{VanWyhe2007,richards2015}: Darwin simply needed more time to gather evidence and develop his argument.

\begin{figure}
\begin{center}
\includegraphics[width=\textwidth]{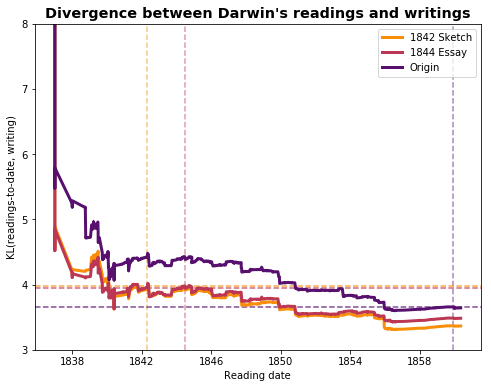}
\end{center}
\caption{\textbf{KL divergence between Darwin's readings and writings} --- Vertical dashed lines indicate date of publication; horizontal dashed lines indicate divergence at the time of publication. \emph{The Origin} diverges more from the readings-to-date than either of the two previous drafts at all time points. However, each successive draft diverges less from the readings-to-date at the time of writing. }
\label{fig:doc-divergence-history}
\end{figure}

We use KL-divergence to trace the increase in cognitive surprise through Darwin's written presentations of his theory. Figure~\ref{fig:doc-divergence-history} shows that with respect to the set of readings at any given time, \emph{The Origin} is significantly more divergent than either of the earlier essays, and that the 1844 essay is slightly more divergent from the readings than the 1842 version. Interestingly, however, the 1842 and 1844 essays are more divergent from Darwin's readings at their respective times of writing, than the Origin is by 1859. This computational evidence supports the claim that Darwin's continued reading during the period between 1844 and 1859 was materially relevant to what he eventually wrote. %but  each iteration of what he wrote is more novel than the previous version. 

%We use Darwin's drafts in 1842 and 1844 to show how his ideas developed into \emph{The Origin}. We compare the average topic distribution of the samples for the Sketch of 1842, Essay of 1844, and \emph{The Origin} to the average topic distribution of Darwin's cumulative readings using the KL divergence. Figure~\ref{fig:doc-divergence-history} shows that for each successive draft (1842, 1844, and \emph{The Origin}), divergence to the readings increases. This new digital evidence shows that the additional time spent reading significantly increased the novelty of \emph{The Origin}, increasing cognitive surprise.

\subsection{The rush to publish: Wallace's essay}
Regardless of the primary cause of Darwin's delay, his sudden rush to publication is often attributed to the co-discovery of natural selection by  Wallace, whose own essay ``On the tendency of varieties to depart infinitely from the original type'' was co-published with an excerpt of Darwin's 1844 essay on 30 June 1858. When Darwin received Wallace's manuscript on 18 June 1858, Darwin had already been organizing his notes for \emph{The Origin} for four years. Writing to Lyell, Darwin remarked on the impressive similarity to his earlier work:
\begin{quote}
I never saw a more striking coincidence. If Wallace had my M.S. sketch written out in 1842 he could not have made a better short abstract! \citep{DCP2285}
\end{quote}
We take Darwin's remark as both praising Wallace's work and emphasizing how much further his own ideas had developed by 1858. Darwin's observation indicated not just similarity between their work, but a \emph{specific} similarity to his 1842 description of natural selection. The JSD measure\footnote{We use the symmetric JSD rather than KL-divergence, as JSD does not make any assumption about ordering of texts.} partially captures Darwin's observation: Wallace's work is more similar to the 1842 and 1844 essays than to \emph{The Origin}. However, it is marginally---just over $1/100th$ of a bit---closer to the 1844 essay than the 1842 essay by this measure (top of Figure~\ref{fig:doc-sims}). Darwin's mention of his 1842 sketch may be interpreted as a generic reference to the earlier period, or it may reflect features of the 1842 sketch not accessible via topic modeling.

\subsection{Dating the ``Outline and Draft''}

\begin{figure}
\begin{center}
\includegraphics[width=0.8\textwidth]{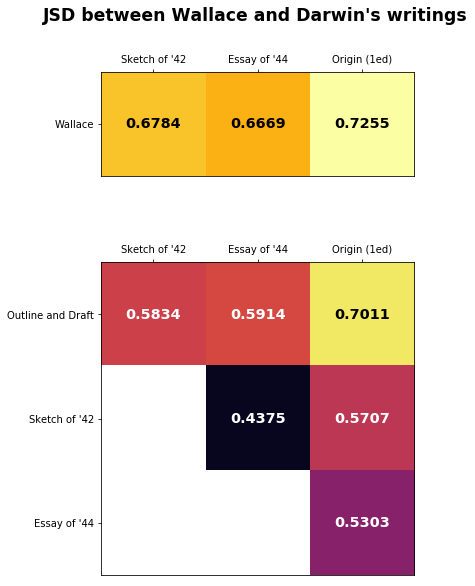}
\end{center}
\caption{\textbf{Similarity between Wallace and Darwin's writings} --- These heatmaps show the Jensen-Shannon Distance (JSD) between Darwin's various writings and Wallace's manuscript.  \emph{Top:} Wallace's text is closer to the two earlier writings than to \emph{The Origin}. \emph{Bottom:} The ``Outline and Draft'' is closest to the Sketch of 1842 and farthest from \emph{The Origin}, indicating it may have been the earliest writing.}
\label{fig:doc-sims}
\end{figure}

Finally, we look at a manuscript originally discovered with the 1842 essay at the Darwin residence in 1896, but which was not included in \emph{The Foundations of the Origin of Species} in 1909\citet{Foundations}. It had fallen into archival obscurity at the Cambridge University Libraries until  rediscovered in 1975 by Peter Vorzimmer,  who dated the outline to July 1839\citet{Vorzimmer1975}.
%when Darwin stopped his ``transmutation'' notebooks and began a three-month work period  in London. 
However, scholarly consensus gravitated to a theory that the paper was an interstitial draft between the 1842 and 1844 essays \citep{Schweber1977,Kohn1982}, based upon annotations on the manuscript itself and the reuse of headings from the draft in the 1844 essay which were not present in the 1842 sketch. Comparing it using JSD, we find that the draft is  further from \emph{The Origin} than either the 1842 sketch or 1844 essay  (bottom of Figure~\ref{fig:doc-sims}). Moreover, the 1842 essay is further from \emph{The Origin} than the 1844 essay. This finding provides some new evidence supporting Vorzimmer’s 1839 dating, although further investigation is necessary.

\end{document}